# Illumination Robust Loop Closure Detection with the Constraint of Pose

Yan Deli[a,b,c*], Tuo Wenkun[b], Wang Weiming[b], and Li Shaohua[a]

[a]State Key Laboratory of Mechanical Behavior and System Safety of Traffic Engineering Structures, Shijiazhuang Tiedao University, Shijiazhuang 050043, China, [b]School of Electrical and Electronic Engineering, Shijiazhuang Tiedao University, Shijiazhuang 050043, China and [c]School of Electronic and Information Engineering, Beijing Jiaotong University, Beijing 100044, China

**Abstract: Background:** Loop closure detection is a crucial part in robot navigation and simultaneous location and mapping (SLAM). Appearance-based loop closure detection still faces many challenges, such as illumination changes, perceptual aliasing and increasing computational complexity.

**Method:** In this paper, we proposed a visual loop-closure detection algorithm which combines illumination robust descriptor DIRD and odometry information. The estimated pose and variance are calculated by the visual inertial odometry (VIO), then the loop closure candidate areas are found based on the distance between images. We use a new distance combing the the Euclidean distance and the Mahalanobis distance and a dynamic threshold to select the loop closure candidate areas. Finally, in loop-closure candidate areas, we do image retrieval with DIRD which is an illumination robust descriptor.

**Results:** The proposed algorithm is evaluated on KITTI_00 and EuRoc datasets. The results show that the loop closure areas could be correctly detected and the time consumption is effectively reduced. We compare it with SeqSLAM algorithm, the proposed algorithm gets better performance on PR-curve.

## 1. INTRODUCTION

SLAM needs to build a map of the current environment based on the collected sensor information and realize self-location in the building map at the same time. Location helps the robot to understand its own state and mapping is an abstract description of the external environment, together helping the robot to perceive the world inside and outside. Loop closure detection is an important part of SLAM which can erase the drift caused by long-term autonomy and understand the real topology of the environment [1].

Visual information is rich and simple to get, so it is widely used in loop closure detection and place recognition [2-4]. A lots of work have been done to help robots better understand the information in the image. Of all the appearance-based SLAM algorithms, FAB-MAP 2.0 which is fully probabilistic and robust against perceptual aliasing is the most successful one. A loop closure detection of the 1000km road network is performed with FAB-MAP 2.0 and shows better recall at 100% precision [5]. SeqSLAM uses a sequence match instead of image match to cope with the extreme perceptual changes [6]. These two methods do loop closure detection in appearance space and all performed well.

Describing a place by visual information is simple and efficient, but it may lead to errors. Images at the same place may vary greatly (affected by lighting, seasons, etc.), and images at different places is sometimes similar. This is a difficult problem for visual loop closure detection known as perceptual aliasing. Furthermore, time to do image retrieval increases rapidly with the growth of trajectory, which makes it no longer applicable in real-time systems. These problems can be solved with the addition of odometry information. With the odometry information, the FAB-MAP and SeqSLAM are improved to become CAT-SLAM and SMART and will be introduced in detail in section 2.

To speed up image retrieval and increase the robustness to illumination and perceptual aliasing, we use the odometry estimated trajectory to constrain the area to do image retrieval and then do image retrieval with DIRD which is robust to illumination. VIO based on extended Kalman filter (EKF) provides the pose and pose variance of each image. We do a preliminary loop closure detection with the pose information and then formed the loop closure candidate area. Image retrieval will then be done on these loop-closure candidate areas to generate the final results. As is shown in Fig. (1), DIRD needs to extract the feature of all images and calculate the similarity of image pairs in the blue trian1gular area (There is a safety margin to skip the pose very near by the current image.). After a preliminary selection, image retrieval only needs to extract the features of involved image and calculate the similarity in loop closure candidate area.

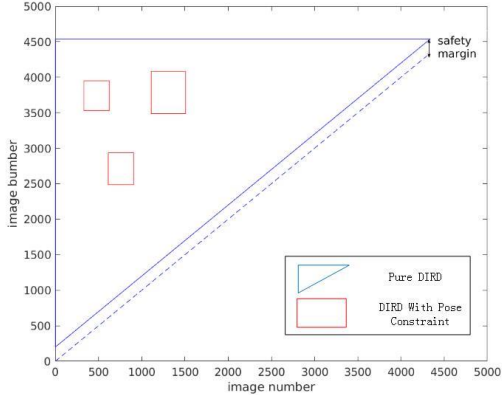

**Fig. (1).** Comparison of image retrieval area. With the constraint of pose, the image retrieval area changes to the red rectangular area from the blue triangular area.

We evaluate the loop closure performance of our proposal on KITTI_00 and EuRoc. These two datasets both provide accurate ground truth and IMU information, so we can get the pose information and extract loop closure. We compare the time consumption with pure DIRD and loop closure results with SeqSLAM. Experimental results shows great decrease in time and a better performance in precision and recall compared with SeqSLAM.

The paper proceeds as follows: Section 2 introduces the development of appearance-based loop-closure detection. Section 3 is background knowledge, including VIO based on EKF and DIRD. Section 4 is our loop-closure detection with the constraint of pose. The experiments and results are shown in Section 5. Section 6 is summary and conclusion.

## 2. RELATED WORKS

With the development of visual SLAM (VSLAM), great achievements have been made in appearance-based place recognition and loop closure detection [7]. The picture description technique is roughly divided into two categories: local feature descriptor and global feature descriptor. Local features selectively extract some image parts and describe them. Typical local descriptors include Scale-Invariant Feature Transforms (SIFT) [8], Speed-Up Robust Features (SURF) [9], BRIEF [10]. Local features in [11-12] are made by combining different descriptors to complete the detection and description. The Bag-of-Words (BOW) model quantifies local features into a vocabulary and then describes the image using a vector implying the availability of a certain local feature in the image. Vocabulary trees can make the detection more efficient, such as the Chou-Liu tree in FAB-MAP in [13]. The global feature directly describes the whole picture scene without a selection phase, such as HOG features [14] and Gist [15]. Local features and global feature descriptions have their own advantages and disadvantages, and they are combined in [16] to overcome some shortcomings when used alone.

In appearance-based loop closure detection, image description is an important part, which has a direct impact on results. SIFT is a big success because it can overcome the influences of rotation and scale changes. But the extraction of SIFT is complex and time-consuming. SURF inherits the advantages of SIFT and speed up the extraction, which has been widely used. Illumination changes have always been a problem in image description, Zambanini et al. constructed a local characterization descriptor that is insensitive to illumination changes in [17], using a Gabor filter and normalizing the filter response. The experimental results show that the descriptor has better performance than the descriptor such as SURF in the case of illumination changes. Lategahn *et al*. in [18-19] presented an image descriptor that is robust to illumination and was named DIRD, and explained the related loop-closure detection algorithm in detail. DIRD consists of the Haar features of the image and can be quantified to bits or bytes to meet different needs. DIRD is tested on multiple data sets. The experimental results show that it has significant advantages when compared with other commonly used descriptors when the illumination changes.

Pure appearance-based loop closure detection is based solely on the image similarity. The time to calculate the similarity increases with growth of image numbers. Much works have been done to shorten the image retrieval time. In the BOW model, researchers used different tree structures in [20-21] to build a dictionary for efficient retrieval. In addition, image retrieval can be accelerated by inverted indices which store the image number against the corresponding word instead of storing the word against the image number. Inverted indices quickly eliminate unlikely images, thereby avoiding traversing the image directly and reducing retrieval time. Mohan et al. divided the dataset into different environments and choose the most likely environment for current image by cooccurent feature matrices in [22]. Through the classification and selection of the environment, the amount of the images that need to be traversed decreases quickly, making the long-term autonomy possible.

The proposals above reduce the time of loop closure detection through different methods. Recent years, much works have been focused on the loop closure detection with image retrieval and topological-metric maps. The topological-metric map can provide positional constraints which means that when performing image retrieval, only the location close to the current location is compared. This not only speeds up the image retrieval, but also enhances the robustness of loop closure detection. Maddern et al. combined odometry information with FAB-MAP and formed CAT-SLAM in [23]. Rao-Blackwellised particle filter is used to achieve the combination of metric information and image visual information. CAT-SLAM is tested under multiple data sets and can detect more correct loops than FAB-MAP. Pepperell et al. combined the SeqSLAM with the camera's own motion information and proposed the SMART algorithm in [24]. The experimental results show that SMART is robust to illumination changes and vehicle speed changes and realizes up to 96% recall rate at 100% accuracy.

## 3. BACKGROUND

In this section, we introduce the visual inertial odometry based on EKF and DIRD which are two essential components of our algorithm. VIO provides the estimated pose and its variance with motion model and observation model. These information is then used to constrain the area of image retrieval. We also introduce the DIRD and its algorithm. DIRD was constructed by a set of elementary algorithmic building blocks and tested to be illumination robust. With VIO based on EKF and DIRD, the image retrieval is more efficient and robust to the illumination.

### 3.1 VIO Based on EKF

The pose and variance are calculated by VIO with the IMU motion information and visual information. The whole process can be divided into two steps: propagation and measurement update.

Consider a SLAM system with the rotation and position of IMU $[R, P_t]$, the velocity $V$ and the position of feature points $[p^1, p^2 \cdots p^m]$. Then the state of the system is described by a vector variables $X_n = [R, V, P_t, p^1, p^2, \cdots p^m] \in \Re^{m+5}$. $Y_m = [y^1, y^2 \cdots y^m]$ is the observation of the feature points. The motion model and the observation model is

$$X_n = f(X_{n-1}, u_n, \omega_n), \qquad (1)$$

$$Y_n = b(X_n) + V_n, \qquad (2)$$

where $f$ is the motion function; control input $u_n$ is the Angular velocity and acceleration of IMU; $\omega_n$ is the noise which is a centered random variable with variance matrix $Q_n$; $b$ is the observation function; $V_n$ is the observation noise.

Firstly, the $\hat{X}_{n|n-1}$ is obtained through motion function

$$\hat{X}_{n|n-1} = f(\hat{X}_{n-1|n-1}, u_n, 0). \qquad (3)$$

The EKF linearize the error system by a first-order Taylor expansion of motion function $f$ and observation function $b$. then the variance of the $\hat{X}_{n|n-1}$ is

$$P_{n|n-1} = F_n P_{n-1|n-1} F_n^T + G_n Q_n G_n^T, \qquad (4)$$

With $F_n = \frac{\partial f}{\partial X}(\hat{X}_{n-1|n-1}, u_n, 0)$, $G_n = \frac{\partial f}{\partial \omega}(\hat{X}_{n-1|n-1}, u_n, 0)$

The next step is measurement update, the $\hat{X}_{n|n}$ and its variance is

$$\hat{X}_{n|n} = \hat{X}_{n|n-1} + K_n z_n, \qquad (5)$$

$$P_{n|n} = (I - K_n H_n) P_{n|n-1}, \qquad (6)$$

where $z_n = Y_n - b(\hat{X}_{n|n-1})$; $K_n$ is the Kalman gain;

$H_n = \frac{\partial b}{\partial X}(\hat{X}_{n-1|n-1})$

### 3.2 DIRD

Fig. (2) shows the construction of DIRD. At first the image is divided into 4*4 segments, then a couple of Haar filters are applied on these segments. The results from the filters of one pixel are put together to form a auxiliary vector. The auxiliary vector is L2 normalized and then be summed over pixel offsets. The results of summation at predefined pixels are concatenated. At last, the concatenated vector is quantified to either bit or byte.

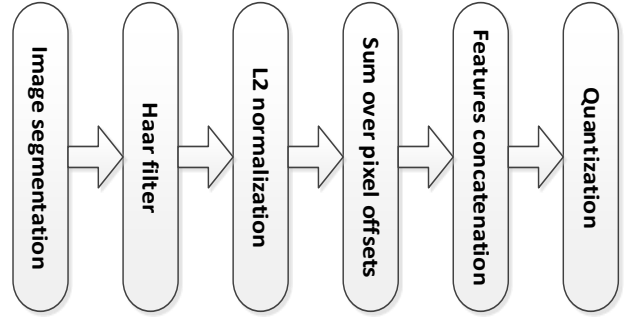

**Fig. (2).** Construction of DIRD descriptor. DIRD is learned by using a set of elementary building blocks from which millions of different descriptors can be constructed automatically.

We use the DIRD (byte) for our detection. A pair of matching images and corresponding DIRD descriptors is shown in Fig. (3). The quantified DIRD (byte) is shown by a gray scale image of 54*64.

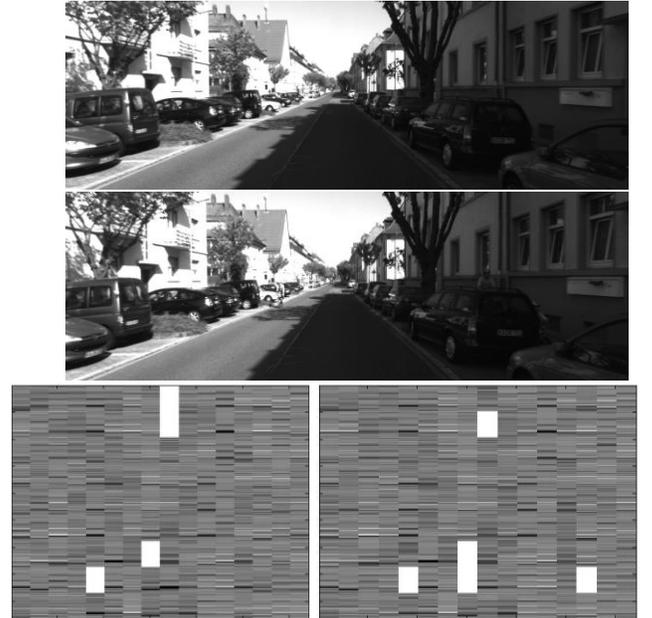

**Fig. (3).** Matching images and corresponding DIRD descriptors. DIRD (byte) is a 3456-dimensional vector with the value in [1,256] and is shown by a gray scale image.

Lategahn et al. perform loop-closure detection on KITTI using DIRD. Their algorithm can be divided into 3 steps including computing features, computing similarity and post processing. At first, the DIRD feature of each image is calculated and saved as feature matrix. Next, the Euclidean distance of two DIRD feature is obtained and translated into similarity by a logistic function. If the similarity is above the threshold, it will be saved to the similarity matrix. Finally, in order to increase the precision of loop-closure, they also do a sequence match and non-maximum suppression. Sequence match requires the sum of the similarities of a sequence is above a threshold. Non-maximum suppression eliminates the false loops around true loops. After the post processing, the similarity matrix only contains only non-zero entries for very likely loop closure.

## 4. LOOP CLOSURE WITH CONSTRAINT OF POSE

### 4.1 Limitations and Values of Position Constraints

Error accumulates with the running of the odometry, eventually leading to draft. Therefore, relying solely on the odometry information to estimate the loop-closure is unreliable, but it does not mean that the odometry estimation information is completely worthless. The odometry information can provide local measurement information and global pose information. The local information can be used as the basis for selecting key frames. The global information may not provide the exact position of the loop-closure, but it can provide loop-closure detection candidate area, thereby reducing image numbers of image retrieval and avoiding perceptual aliasing.

Fig. (4) illustrates the process of odometry estimation, where the black triangle represents the estimated position of the robot, represented by a vector $\hat{X}_k$, and the red dotted triangle represents the true position, represented by a vector $X_k$. For a clear representation, the diagram only shows the pose deviations for the first three moments of the loop-closure. The nearest neighbor method is used to estimate the loop-closure and it's important to choose a proper threshold $\lambda$. When the pose deviation is smaller than $\lambda$, there is a possibility of loop-closure. When it is greater than $\lambda$, it is considered as a new position. As the error accumulates, some loop-closure area may not be found, so that the loop-closure can't be effectively detected. In the estimation process, the pose variance describes the uncertainty of the estimated pose, so it is considered to add the pose variance to the distance calculation, instead of relying on the Euclidean distance for the loop-closure area judgment.

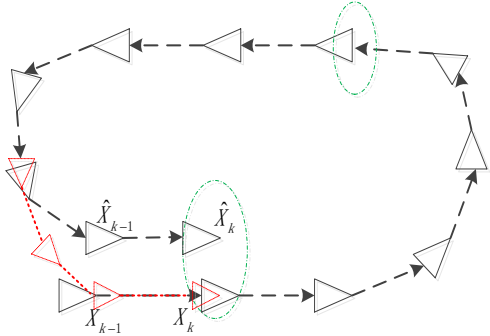

**Fig. (4).** Loop-closure Detection with pose constraint. Estimated trajectory deviates from actual position due to cumulative error. Using a proper threshold, loop-closure can also be found but may not be that accurate.

### 4.2 Selection of loop-closure candidate area

With the pose information from VIO, now the description of the place is composed of DIRD feature and pose information. The place description is

$$U_k = \{T_c(k), D(k)\}, \tag{7}$$

where $D(k)$ is DIRD feature; $T_c(k)$ is composed of position ${}^Wp_c(k)$, rotation ${}^WR_c(k)$, position variance $P_p(k)$ and rotation variance $P_R(k)$.

The pose estimation become unreliable after a long time running, so using the Euclidean distance between two poses can't provide us with meaningful loop-closure. The Mahalanobis distance which contains the variance information can effectively calculate the similarity between the two sample sets. However, Mahalanobis distance is quite sensitive to the changes of variance, so we use a distance combining the Euclidean distance and the Mahalanobis distance and it is

$$d(U_i, U_j) = \sqrt{\Delta p_{i,j}^T * P_{i,j}^{-1} * \Delta p_{i,j}}, \tag{8}$$

where the $\Delta p_{i,j}$ and $P_{i,j}$ is

$$\begin{cases} \Delta p_{i,j} = {}^Wp_c^i - {}^Wp_c^j \\ P_{i,j} = I + P_p^i + P_p^j \end{cases} \tag{9}$$

This distance is similar to the Mahalanobis distance definition, while the variance representing the uncertainty of the distance is the corrected variance as defined above instead of the position covariance at two moments. The variance can reduce the influence of the uncertainty of the distance and reasonably reflect the spatial position distance. When the variance is 0, the distance becomes the Euclidean distance. When the variance grows, the distance will be smaller than Euclidean distance, reducing the influence of the uncertainty of the distance.

With the definition of the distance, computers efficiently compute the distance between images and find out the nearest one of current image. Then we need a threshold to judge whether they can be accepted as a loop-closure. Using EKF, the error of the variable satisfies $e_{n|n} \sim N(0, P_{n|n})$. For moment i, $P(|X_i - \hat{X}_{i|i}| < 1.96\sqrt{P_p^i}) \approx 0.95$, so we want the distance is less than the radius of possible appearance area which is at 95% confidence level. Then the distance should satisfy:

$$\min_{j=0..i-30} \{d(U_i, U_j)\} < 1.96 | \sqrt{P_p^i} | + \beta \tag{10}$$

We also add an extra $\beta$ which is a reasonable drift during the movement from j to i. In our experiments, we set the $\beta$ as a constant according to the environment scale.

Using odometry pose information and the distance above, we have the preliminary results of loop-closure. Judging by the continuity, the loop-closure results is divided into different cluster and form different loop-closure candidate areas which was shown in Experiments. We perform DIRD in these area and use this as the final results of loop-closure.

## 5. EXPERIMENTS

We evaluate our algorithm on two public datasets: EuRoc [27] MH_05_difficult and KITTI_00. The KITTI_00 is captured from a moving vehicle driving through the city of Karlsruhe. EuRoc was collected by a micro aerial vehicle in an industrial environment. Both datasets provide accurate ground truth. The trajectory of the datasets is plotted and shown in green in Fig. (5). Judging from the distance between each pose and its nearest neighbor, we get the true loop closure and is shown in red in Fig. (5). This loop closure results serve as reference for the precision and recall. All of the experiments were performed on the computer with an Intel Core i5-5200U CPU with 2.60 GHz, and 8 GB RAM.

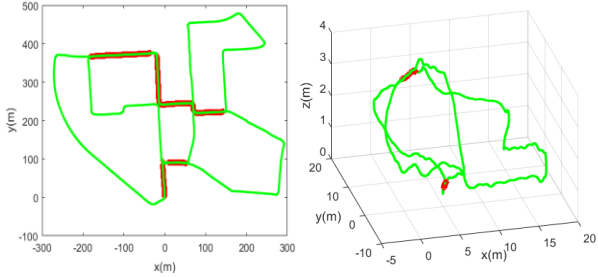

**Fig. (5).** The trajectory of the KITTI (left) and EuRoc (right), the red parts of the trajectory is the true loop-closure using ground truth.

### 5.1 KITTI

At first, loop-closure is obtained using the pose information from odometry. Due to the cumulative error, these loops may not be accurate enough. Fig. (6) shows the difference between true loop-closure and loop-closure obtained using odometry pose information. We cluster the preliminary loop-closure results according to its contnuity and surround each cluster with a rectangle. These rectangular areas are what we called loop-closure candidate area. The rectangles is properly enlarged to avoid the missing of true loops at both ends.

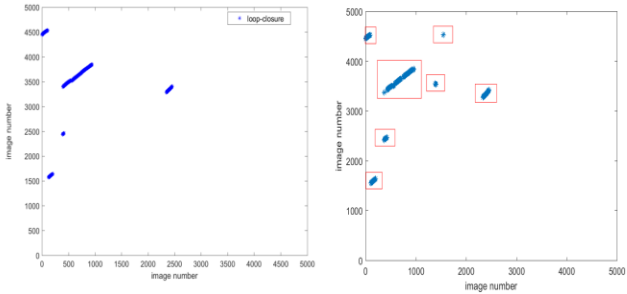

**Fig. (6).** The corresponding image number of loop-closure is depicted,true loop-closure (left) and loop-closure obtained using odometry pose information(right) is different. The red rectangles are the loop-closure candidate areas.

Then we perform image retrieval in these areas. Compared with traversing the whole images, image retrieval on these areas is much time-saving. The time to extract the image feature will also decrease because there is no need to extract features of all images. The time to extract DIRD feature for each image is depicted in Fig. (7).

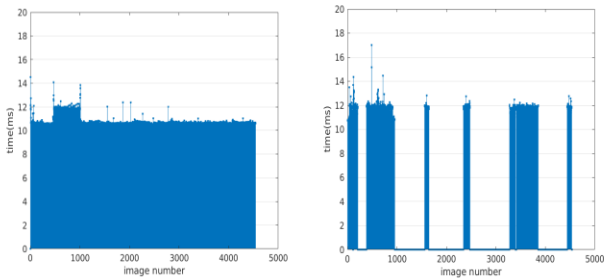

**Fig. (7).** The time to extract image feature. Time to extract features in loop-closure candidate area(right) is shorter than to extract features of all images.

The total time of loop-closure detection is shown in Fig. (8). Pure DIRD takes 182763.66ms to traverse the whole dataset compared with 28329.93ms when adding the constraint of pose. The time to calculate similarity doesn't increase with the growth of image number. It is influenced by the horizontal length of the loop-closure candidate area.

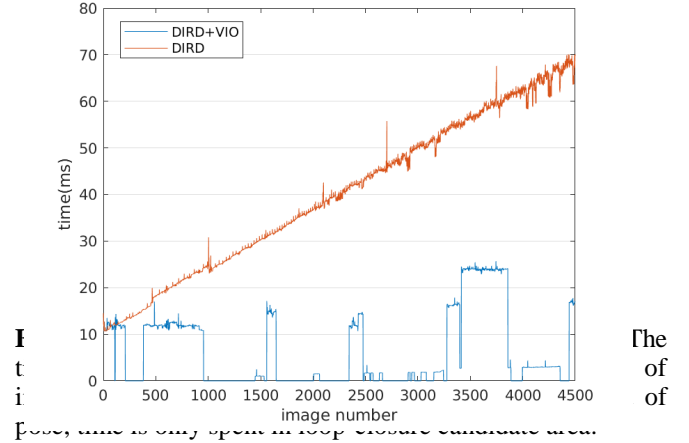

**Fig. (8).** The total time of loop-closure detection. The time of DIRD is increasing while adding the constraint of pose, time is only spent in loop-closure candidate area.

Then we compare our loop-closure results with SeqSLAM through PR-curve. The loop-closure results and PR-curve is shown in Fig. (9). After performing DIRD in loop-closure candidate area, some incorrect loops are removed and the loop-closure is more accurate.

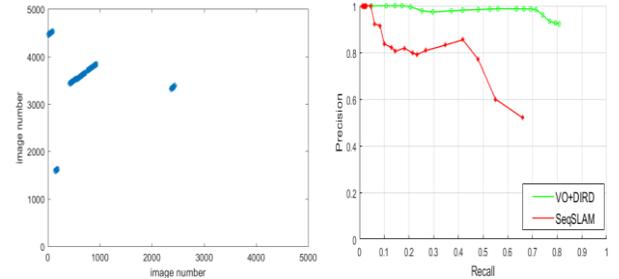

**Fig. (9).** The results of loop-closure detection (left) and the PR-curve compared with SeqSLAM (right). DIRD with pose constraint shows higher precision at the same recall.

### 5.2 Euroc

The trajectory of Euroc is shown above in Fig. (4). Compared with KITTI, its trajectory is 3-dimensional and the loop-closure is less. So, it's harder to get an accurate trajectory through VIO. Fig. (10) (left) shows the estimated trajectory and loop-closure results obtained using the pose information.To show the difference between the loop-closure results, loop-closure using ground truth and pose information from VIO is pictured together in Fig. (10) (right). Although the loop-closure results using pose information is not accurate enough, the true loops is in the vicinity of it. Then the loop-closure candidate area is given according to the continuity and the true loop is well surrounded by it.

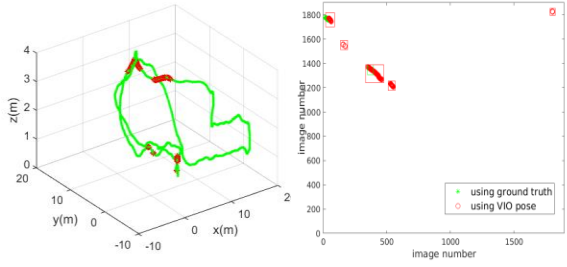

**Fig. (10).** Estimated trajectory and loop-closure obtained using odometry pose information (left). Comparison of true loop-closure and loop-closure using odometry pose information (right).

As is shown above in Fig. (8), the time grows continuously when DIRD used alone while the time is stable in loop-closure candidate area. In fact when detecting loop-closure using pose information, there is also a increase in time with the growth of image number. But pose comparison is relatively faster and we only cares about the nearest neighbor. So even there is an extra pose comparison, time still decrease with the constraint of pose. Loop-closure detection time of Euroc dataset is shown in Fig. (11).

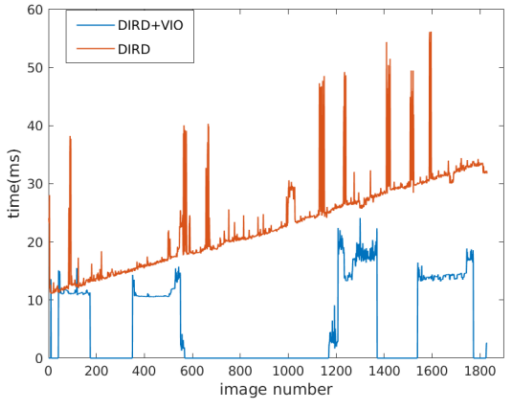

**Fig. (11).** Time consumption on each image. With the constraint of pose, a lots of feature extraction time is saved.

At last, we compare the results with SeqSLAM through PR-curve again and it still shows a higher precision at the same recall. The results is shown in Fig. (12).

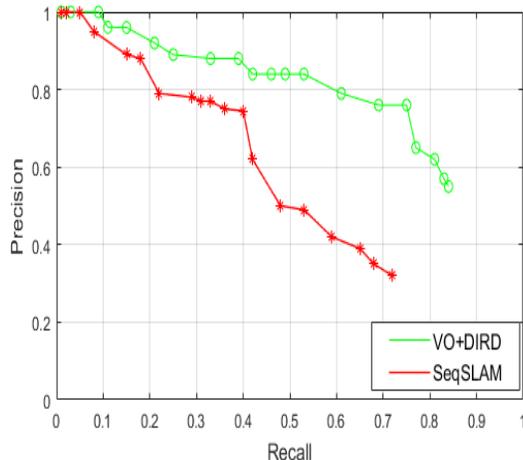

**Fig. (12).** PR-curve of Euroc compared with SeqSLAM,

there is a decrease in precision due to the complexity of the environment compared with KITTI.

## 6. CONCLUSION AND SUMMARY

Loop-closure detection is an important part of SLAM. The cumulative error of pose estimation can be effectively eliminated with loop-closure detection. Aiming at the problem of perception aliasing, illumination variation and the increase of computational complexity in large scenes during visual loop-closure detection, we introduce the pose information from VIO to moderate these problems. With the pose constraint, loop-closure detection time decreases and loop-closure results is more accurate.

In this paper, we present our loop closure detection combing DIRD and VIO pose information. Our loop-closure detection is mainly about two stages. Stage one, preliminary results of loop-closure detection is obtained using the pose information from VIO. We used a distance combining the the Euclidean distance and the Mahalanobis distance and clustered the results to form loop-closure candidate area. Stage two, we perform DIRD in loop-closure candidate area to get the final results. DIRD is robust to illumination changes and efficiently remove the incorrect loops. We do experiments on KITTI and EUROC, both experiments show better precision at the same recall and a decrease of time consumption.

## REFERENCES

[1] Cadena C , Carlone L , Carrillo H , et al., "Past, Present, and Future of Simultaneous Localization and Mapping: Toward the Robust-Perception Age", IEEE Transactions on Robotics, 2016, 32(6):1309-1332.

[2] F. Chaumette, P. Corke, and P. Newman, "Editorial Special Issue on Robotic Vision", The International Journal of Robotics Research, vol. 29, pp. 131-132, 2010.

[3] G. Hager, M. Hebert, and S. Hutchinson, "Editorial: Special Issue on Vision and Robotics, Parts I and II", International Journal of Computer Vision, vol. 74, pp. 217-218, 2007.

[4] J. Neira, A. J. Davison, and J. J. Leonard, "Guest Editorial Special Issue on Visual SLAM," IEEE Transactions on Robotics, vol. 24, pp. 929-931, 2008.

[5] M. Cummins and P. Newman,"Appearance-only SLAM at large scale with FAB-MAP 2.0", Int. J. Robot. Res., vol. 30, no. 9, pp. 1100–1123, 2011.

[6] Milford M J , Wyeth G F, "SeqSLAM: Visual route-based navigation for sunny summer days and stormy winter nights", Robotics and Automation (ICRA), 2012 IEEE International Conference on. IEEE, 2012.

[7] S. Lowry et al.,"Visual place recognition: A survey",IEEE Trans. Robot.,vol. 32, no. 1, pp. 1–19, Feb. 2016.

[8] D. Lowe,"Object recognition from local scale-invariant features,"in Proc. IEEE Int. Conf. Comput. Vis., 1999, vol. 2, pp. 1150–1157.

[9] H. Bay, T. Tuytelaars, and L. Van Gool,"SURF: Speeded up robust features",in Proc. Eur. Conf. Comput. Vis., 2006, pp. 404–417.

10] M. Calonder, V. Lepetit, M. Ozuysal, T. Trzcinski, C. Strecha, and P.Fua,"BRIEF: Computing a local binary descriptor very fast",IEEE Trans. Pattern Anal. Mach. Intell., vol. 34, no. 7, pp. 1281–1298, Jul.2012.

[11] C. Mei, G. Sibley, M. Cummins, P. Newman, and I. Reid,"A constant-time efficient stereo SLAM system",presented at the Brit. Mach. Vis. Conf., London, U.K., 2009.

[12] W. Churchill and P. Newman,"Experience-based navigation for long-term localisation", Int. J. Robot. Res., vol. 32, no. 14, pp. 1645–1661,2013

[13] Cummins M , Newman P, "FAB-MAP: Probabilistic Localization and Mapping in the Space of Appearance", Sage Publications, Inc. 2008.

[14] HE N,CAO J,SONG L, "Scale space histogram of oriented gradients for human detection", International Symposium on Information Science and Engineering.Shanghai:IEEE,2008:167-170.

[15] A. Oliva and A. Torralba,"Building the gist of a scene: The role of

global image features in recognition", in Visual Perception-Fundamentals of Awareness: Multi-Sensory Integration and High-Order Perception. New York, NY, USA: Elsevier, 2006, pp. 23–36.

[16] A. Murillo, G. Singh, J. Kosecka, and J. Guerrero,"Localization in urban environments using a panoramic gist descriptor",IEEE Trans. Robot.,vol. 29, no. 1, pp. 146–160, Feb. 2013

[17] Zambanini S , Kampel M , "A Local Image Descriptor Robust to Illumination Changes",Scandinavian Conference on Image Analysis. 2013.

[18] Lategahn H, Beck J, Stiller C, "DIRD is an illumination robust descriptor", IEEE Intelligent Vehicles Symposium. 2014.

[19] Lategahn H , Beck J , Kitt B , et al., "How to Learn an Illumination Robust Image Feature for Place Recognition", IEEE Intelligent Vehicles Symposium. IEEE, 2013.

[20] Galvez-LoPez D , Tardos J D, "Bags of Binary Words for Fast Place Recognition in Image Sequences", IEEE Transactions on Robotics, 2012, 28(5):1188-1197.

[21] D. Nister and H. Stewenius, "Scalable recognition with a vocabulary tree", in Proc. IEEE Comput. Soc. Conf. Comput. Vis. Pattern Recog., 2006, vol. 2, pp. 2161–2168.

[22] M. Mohan, D. Galvez-Lopez, C. Monteleoni, and G. Sibley, "Environment selection and hierarchical place recognition", in Proc. IEEE Int.Conf. Robot. Autom., May 2015, pp. 5487 –5494.

[23] Maddern W , Milford M , Wyeth G, "CAT-SLAM: probabilistic localisation and mapping using a continuous appearance-based trajectory", The International Journal of Robotics Research, 2012, 31(4):429-451.

[24] Pepperell E , Corke P I , Milford M J, "All-environment visual place recognition with SMART", 2014 IEEE International Conference on Robotics and Automation (ICRA).,IEEE, 2014.

[25] Shi J, Tomasi, "Good features to track", IEEE Computer Society Conference on Computer Vision and Pattern Recognition. USA, IEEE: 1994:593–600.

[26] Barrau A , Bonnabel, Silvère, "Invariant Kalman Filtering", 2018 International Conference on Information Fusion (FUSION). 2018:annurev-control-060117-105010.

[27] Burri, M. , et al, "The EuRoC micro aerial vehicle datasets", *The International Journal of Robotics Research*(2016):0278364915620033.